\documentclass[sigconf,authorversion,nonacm]{acmart}
\usepackage{enumitem}
\usepackage{balance}
\AtBeginDocument{%
  \providecommand\BibTeX{{%
    \normalfont B\kern-0.5em{\scshape i\kern-0.25em b}\kern-0.8em\TeX}}}

\copyrightyear{2023}
\acmYear{2023}
\setcopyright{acmlicensed}
\acmConference[MM '23] {Proceedings of the 31st ACM International Conference on Multimedia}{October 29--November 3, 2023}{Ottawa, ON, Canada.}
\acmBooktitle{Proceedings of the 31st ACM International Conference on Multimedia (MM '23), October 29--November 3, 2023, Ottawa, ON, Canada}
\acmPrice{15.00}
\acmISBN{979-8-4007-0108-5/23/10}
\acmDOI{10.1145/3581783.3612425}
\begin{document}

%%
%% The "title" command has an optional parameter,
%% allowing the author to define a "short title" to be used in page headers.
%\title{TikTalk: A Multi-Modal Dialogue Dataset for Real-World Chitchat}
\title{TikTalk: A Video-Based Dialogue Dataset for Multi-Modal Chitchat in Real World}

%%
%% The "author" command and its associated commands are used to define
%% the authors and their affiliations.
%% Of note is the shared affiliation of the first two authors, and the
%% "authornote" and "authornotemark" commands
%% used to denote shared contribution to the research.
\author{Hongpeng Lin}
\authornote{Equal contribution.}
\email{hopelin@ruc.edu.cn}
\orcid{0009-0000-4742-4247}
\affiliation{%
  \institution{Renmin University of China}
  \city{}
  \country{}
  }

\author{Ludan Ruan}
\email{ruanld@ruc.edu.cn}
\orcid{0009-0009-1039-4940}
\authornotemark[1]
\affiliation{%
  \institution{Renmin University of China}
  \city{}
  \country{}
  }

\author{Wenke Xia}
\email{xwk529086017@gmail.com}
\orcid{0009-0000-1597-9512}
\authornotemark[1]
\affiliation{%
  \institution{Renmin University of China}
  \city{}
  \country{}
  }

\author{Peiyu Liu}
\email{liupeiyustu@ruc.edu.cn}
\orcid{0000-0002-2974-9184}
\affiliation{%
  \institution{Renmin University of China}
  \city{}
  \country{}
  }

\author{Jingyuan Wen}
\email{wenjingyuan@ruc.edu.cn}
\orcid{0009-0008-1875-961X}
\affiliation{%
  \institution{Renmin University of China}
  \city{}
  \country{}
  }

\author{Yixin Xu}
\email{xu_yixin@ruc.edu.cn}
\orcid{0009-0007-8315-6842}
\affiliation{%
  \institution{Renmin University of China}
  \city{}
  \country{}
  }

\author{Di Hu}
\email{dihu@ruc.edu.cn}
\orcid{0000-0002-7118-6733}
\affiliation{%
  \institution{Renmin University of China}
  \city{}
  \country{}
  }

\author{Ruihua Song}
\email{songruihua_bloon@outlook.com}
\orcid{0000-0002-2163-7401}
\authornote{Corresponding author.}
\affiliation{%
  \institution{Renmin University of China}
  \city{}
  \country{}
  }

\author{Wayne Xin Zhao}
\email{batmanfly@gmail.com}
\orcid{0000-0002-8333-6196}
\affiliation{%
  \institution{Renmin University of China}
  \city{}
  \country{}
  }

\author{Qin Jin}
\email{qjin@ruc.edu.cn}
\orcid{0000-0001-6486-6020}
\affiliation{%
  \institution{Renmin University of China}
  \city{}
  \country{}
  }

\author{Zhiwu Lu}
\email{luzhiwu@ruc.edu.cn}
\orcid{0000-0003-0280-7724}
\affiliation{%
  \institution{Renmin University of China}
  \city{}
  \country{}
  }
\renewcommand{\shortauthors}{Hongpeng Lin et al.}

%% No italics and no comma

%% If needed use a foot or author note to identify equal contribution
%%
%% By default, the full list of authors will be used in the page
%% headers. Often, this list is too long, and will overlap
%% other information printed in the page headers. This command allows
%% the author to define a more concise list
%% of authors' names for this purpose.
% \renewcommand{\shortauthors}{Anonymous Author, et al.}

%%
%% The abstract is a short summary of the work to be presented in the
%% article.
\begin{abstract}
To facilitate the research on intelligent and human-like chatbots with multi-modal context, we introduce a new video-based multi-modal dialogue dataset, called TikTalk. We collect 38K videos from a popular video-sharing platform, along with 367K conversations posted by users beneath them. Users engage in spontaneous conversations based on their multi-modal experiences from watching videos, which helps recreate real-world chitchat context. Compared to previous multi-modal dialogue datasets, the richer context types in TikTalk lead to more diverse conversations, but also increase the difficulty in capturing human interests from intricate multi-modal information to generate personalized responses. Moreover, external knowledge is more frequently evoked in our dataset. These facts reveal new challenges for multi-modal dialogue models. We quantitatively demonstrate the characteristics of TikTalk, propose a video-based multi-modal chitchat task, and evaluate several dialogue baselines. Experimental results indicate that the models incorporating large language models (LLM) can generate more diverse responses, while the model utilizing knowledge graphs to introduce external knowledge performs the best overall. Furthermore, no existing model can solve all the above challenges well. There is still a large room for future improvements, even for LLM with visual extensions. Our dataset is available at \url{https://ruc-aimind.github.io/projects/TikTalk/}.

\end{abstract}

%%
%% The code below is generated by the tool at http://dl.acm.org/ccs.cfm.
%% Please copy and paste the code instead of the example below.
%%
% \begin{CCSXML}
% <ccs2012>
% <concept>
% <concept_id>10010147.10010178.10010179.10010181</concept_id>
% <concept_desc>Computing methodologies~Discourse, dialogue and pragmatics</concept_desc>
% <concept_significance>500</concept_significance>
% </concept>
% <concept>
% <concept_id>10010147.10010178.10010179.10010182</concept_id>
% <concept_desc>Computing methodologies~Natural language generation</concept_desc>
% <concept_significance>500</concept_significance>
% </concept>
% <concept>
% <concept_id>10010147.10010178.10010224</concept_id>
% <concept_desc>Computing methodologies~Computer vision</concept_desc>
% <concept_significance>500</concept_significance>
% </concept>
% </ccs2012>
% \end{CCSXML}

% \ccsdesc[500]{Computing methodologies~Discourse, dialogue and pragmatics}
% \ccsdesc[500]{Computing methodologies~Natural language generation}
% \ccsdesc[500]{Computing methodologies~Computer vision}

%%
%% Keywords. The author(s) should pick words that accurately describe
%% the work being presented. Separate the keywords with commas.
\keywords{Multi-modal Dialogue; Chitchat; Real world; Dataset}

%% A "teaser" image appears between the author and affiliation
%% information and the body of the document, and typically spans the
%% page.
% \begin{teaserfigure}
%   \includegraphics[width=\textwidth]{sampleteaser}
%   \caption{Seattle Mariners at Spring Training, 2010.}
%   \Description{Enjoying the baseball game from the third-base
%   seats. Ichiro Suzuki preparing to bat.}
%   \label{fig:teaser}
% \end{teaserfigure}

% \received{20 February 2007}
% \received[revised]{12 March 2009}
% \received[accepted]{5 June 2009}

%%
%% This command processes the author and affiliation and title
%% information and builds the first part of the formatted document.
\maketitle

\section{Introduction}

\begin{figure}
    \centering
    \includegraphics[width=0.95\linewidth]{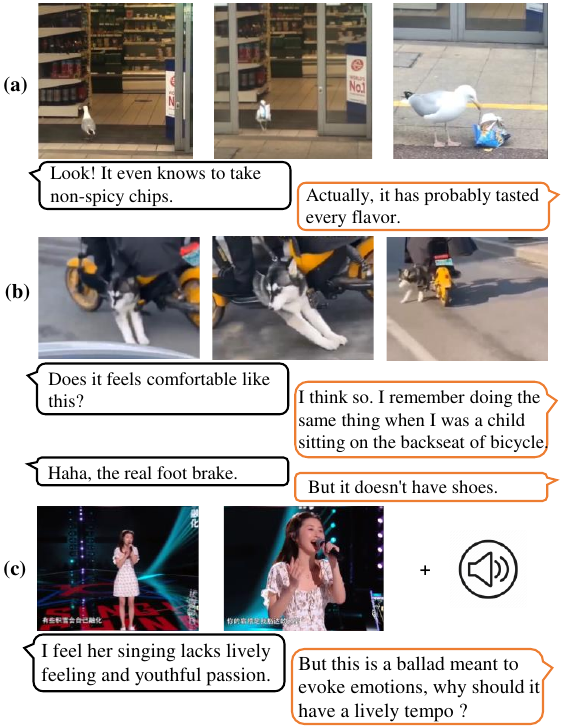}
    \vspace{-0.2cm}
    \caption{Dialogue examples from our TikTalk dataset. (a) A seagull carrying food from a store. (b) A dog intentionally stretches out its paw to rub against the ground while its owner rides an electric bike. (c) A singer is singing a ballad.}
    \vspace{-0.4cm}
    \label{fig:tiktalk_example}
\end{figure}

The ability to perceive and interact with multi-modal information, such as vision and audio, is a significant manifestation of conversational robots’ progress toward artificial general intelligence. Recently, with the excellent performance of conversation models~\cite{zhang2019dialogpt, bao2022plato, adiwardana2020towards, roller2021recipes, zhou2021eva} and large language models~\cite{chatgpt, wang2021ernie, hoffmann2022training, zhang2022opt, chung2022scaling, touvron2023llama}, there have been widespread interests in introducing multi-modal information into conversations. GPT-4~\cite{openai2023gpt4} and some other works~\cite{alayrac2022flamingo, li2023blip} have extended text-based dialogue agents to support both textual and visual inputs. They provide surprisingly excellent results when prompted with questions related to visual content. However, current ways to ask questions about visual content ~\cite{das2017visual, alamri2019audio} are just designed to test the model's abilities in objective perception and simple comprehension of visual information, and they still differ largely from human daily dialogues with multi-modal context.

To facilitate the research on multi-modal dialogue, various forms of datasets have been proposed. IGC~\cite{mostafazadeh2017image}, Image-Chat~\cite{shuster2020image} and PhotoChat~\cite{zang2021photochat} construct dialogues around given images via crowdsourcing. MMDD~\cite{lee2021constructing} and DialogCC~\cite{lee2022dialogcc} match textual dialogue datasets with image caption datasets. MMChat~\cite{zheng2022mmchat} and MMDialog~\cite{feng2022mmdialog} are derived from image posts and comment-reply pairs on social media. 
OpenViDial~\cite{meng2020openvidial, wang2021openvidial} and M$^3$ED~\cite{zhao2022m3ed} extract dialogues of characters and corresponding scenes from movies and TV series.

However, there still exist certain gaps between these previous datasets and real-world multi-modal dialogues. Firstly, people engage in spontaneous dialogues under their shared experiences of multi-modal information from the real world, whereas most multi-modal dialogue datasets do not simulate this scenario. For example, the dialogues collected via crowdsourcing follow explicit guidelines and lack spontaneity and authenticity, leading to limited quality and coverage. Textual dialogue and image caption datasets for multi-modal dialogues are not naturally aligned. Additionally, character dialogues in movies and TV series are designed to meet the demands of storyline, and may be manipulated for dramatic effects. Secondly, real-world multi-modal dialogue scenarios involve various context, enabling individuals to engage in personalized and diverse dialogues. Most of the existing datasets provide limited visual information from static images. Furthermore, the specific environmental context in movies and TV series, as well as the predominant visual content of the dialogue scenes being the talking characters, may also restrict the diversity of conversations.

In this paper, we aim to introduce a novel video-based dialogue dataset that aligns with multi-modal chitchat scenarios in real world.
Firstly, we construct a Chinese multi-modal dialogue dataset called TikTalk by collecting videos and their corresponding comments and replies from a video-sharing platform. TikTalk consists of 38K videos and 367K dialogues. Figure~\ref{fig:tiktalk_example} shows several examples from TikTalk, where users engage in spontaneous dialogues after watching videos, offering a better simulation of real-world chitchat context. These dialogues involve a diverse range of multi-modal context. For example, the responses in Figure~\ref{fig:tiktalk_example} (a) and (c) are derived from visual and audio content, respectively. In Figure~\ref{fig:tiktalk_example} (b), the dialogue requires external knowledge, including personal childhood experiences and the association between the concept of ``brake'' and the action of rubbing against the ground. Moreover, users may capture different interests from the same context, as depicted in Figure~\ref{fig:tiktalk_example} (b), where one user cares for whether the dog is comfortable, while the other notices that it does not wear shoes while rubbing against the ground.
Secondly, we quantitatively compare our dataset with previous ones from the new perspective of which type of context is necessary when generating responses. Results show that the responses in TikTalk require a more diverse range of context. Additionally, we define a multi-modal chitchat task based on TikTalk and summarize three new challenges: (1) Perceiving more diverse multi-modal context and comprehending their complex interaction; (2) Capturing human interests from rich information to generate the desired responses; (3) Introducing external knowledge related to the video as supplementary context.
Thirdly, we evaluate some dialogue baselines on TikTalk, including two mainstream architectures: (1) Using adapter to bridge the semantic gap between vision encoder and language language model; (2) Directly fusing multi-modal features. We find that introducing visual information and external knowledge can improve the quality of responses. Regarding the use of knowledge to generate responses, large language models performs worse than knowledge graphs. However, existing baselines still have some issues, such as failing to utilize audio information effectively and a lack of exploration of the second challenge.

There are three main contributions of our paper: 
\begin{enumerate}
    \item We construct a new video-based dialogue dataset called TikTalk, simulating real-world multi-modal chitchat scenarios;
    \item   We quantitatively demonstrate the characteristics of TikTalk, propose a video-based multi-modal chitchat task and summarize three new challenges brought by TikTalk;
    \item We evaluate several baselines in two mainstream frameworks. Experimental results demonstrate that there is still a lot of room for future improvements on TikTalk.
\end{enumerate}

\begin{table*}[!ht]
    \centering
    \caption{A summary of main multi-modal dialogue datasets and their characteristics. The dialogue types can be classified into three categories, including question-answer dialogues(Q\&A), dialogues that take place in multi-modal environments(in-scene), and dialogues based on multi-modal information(multi-modal based). Modalities include vision(v), text(t), and audio(a).}
    \vspace{-0.25cm}
    \setlength{\tabcolsep}{2.7mm}{
    \begin{tabular}{ccccccc}
    \hline
         \textbf{Dataset} & \textbf{Dialogue Type} & \textbf{Modalities}   & \textbf{Dialogue Source} & \textbf{Turns}  & \textbf{Language} & \textbf{Vision Source} \\
         \hline
         VisDial~\cite{das2017visual} &Q\&A& v, t & crowdsourcing & 2.47M & English & MS COCO \\
         \hline
         AVSD~\cite{alamri2019audio} & Q\&A&a, v, t & crowdsourcing & 236K & English & Charades \\
         \hline
         IGC~\cite{mostafazadeh2017image}& multi-modal based &  v, t & crowdsourcing & 25.3K& English&VQG\\
         \hline
         Image-Chat~\cite{shuster2020image}& multi-modal based & v, t & crowdsourcing & 401K & English & YFCC100M\\
         \hline
         PhotoChat~\cite{zang2021photochat}& multi-modal based & v, t & crowdsourcing & 156K & English & OpenImage V4 \\
         \hline
         MMDD~\cite{lee2021constructing}& multi-modal based & v, t & text datasets & 346K & English & MS COCO, Flicker30K\\
         \hline
         DialogCC~\cite{lee2022dialogcc}& multi-modal based & v, t & text datasets & 929K & English & CC3M\\
         \hline
         MMChat~\cite{zheng2022mmchat}& multi-modal based & v, t & social media &314K & Chinese & Weibo\\
         \hline
         MMDialog~\cite{feng2022mmdialog}& multi-modal based & v, t & social media & 4.92M  & English & Social Media\\
         \hline
         OpenViDial~\cite{meng2020openvidial}& in-scene & v, t & movies\&TVs & 1.1M  & English & Movies\&TVs\\
         \hline
         OpenViDial 2.0 ~\cite{wang2021openvidial} & in-scene& v, t & movies\&TVs & 5.6M  & English & Movies\&TVs\\
         \hline
         YTD-18M~\cite{han2023champagne} & in-scene& a, v, t & transcripts & 54M & English & YouTube\\
         \hline
         MLED~\cite{poria2019meld} & in-scene& a, v, t & TVs & 13.7K & English & TVs\\
         \hline
         M$^{3}$ED~\cite{zhao2022m3ed} & in-scene& a, v, t & TVs & 24.4K & Chinese & TVs\\
         \hline
         \textbf{TikTalk(Ours)}& multi-modal based & a, v, t & social media & 827K & Chinese & Douyin\\
         \hline
    \end{tabular}}
    \vspace{-0.3cm}
    \label{tab:dataset_comparison}
\end{table*}
\section{Related Work}
\label{sec:related}

\subsection{Multi-modal Dialogue Datasets}
Dialogue has been a subject of research interest for NLP community for many years. Many textual dialogue datasets are available~\cite{sordoni2015neural, lison2016opensubtitles2016, wu2017sequential, zhang2018personalizing, dziri2019augmenting, wang2020large}. There have been attempts to introduce multi-modal information into dialogue datasets. We summarize main multi-modal dialogue datasets and their characteristics in Table~\ref{tab:dataset_comparison}.

Some datasets ~\cite{das2017visual,de2017guesswhat, alamri2019audio} aim to address multi-turn Q\&A tasks. They are created with the purpose of measuring models performance on objective perception for given multi-modal information.
Some studies use crowdsourcing, textual dialogue datasets, or collect comment-reply pairs from social media to construct image-based dialogue corpora. IGC~\cite{mostafazadeh2017image} and Image-Chat~\cite{shuster2020image} ask crowd-workers to talk about given images and related topics with explicit guidelines. In most cases, these utterances are highly image-dependent. PhotoChat~\cite{zang2021photochat} simulates a conversation scenario of photo sharing. MMDD~\cite{lee2021constructing} and DialogCC~\cite{lee2022dialogcc} attempt to align textual dialogue datasets and image caption datasets. MMChat~\cite{zheng2022mmchat} and MMDialog~\cite{feng2022mmdialog} collect images and dialogues from social media. MMDialog may contain images in both context and responses. Compared to these works, our dialogues are based on videos containing rich temporal multi-modal information, rather than static images.

Researchers have also made attempts at constructing 
video-based dialogue datasets from different perspectives. OpenViDial datasets~\cite{meng2020openvidial, wang2021openvidial} extract dialogue between characters from movies and TV series and use corresponding screenshots as vision context. The characters' lines are designed to drive plots and have artistic qualities. YTD-18M~\cite{han2023champagne} crawls online videos and constructs dialogues through transcripts. However, the dialogue between characters in video scenes still accounts for the majority of the data, and converting transcripts into dialogues may lead to inaccuracies. MLED~\cite{poria2019meld} and M$^3$ED~\cite{zhao2022m3ed} are designed to predict the emotions of dialogues in TV series. Live comment generation~\cite{ma2019livebot, wang2020videoic} aims to generate comments about a specific moment in a video or to respond to other comments. However, the comments often lack clear context and contain many redundant corpora. 

Our TikTalk dataset comprises of comment-reply dialogues collected from a video-sharing platform. Users engage in spontaneous chitchat based on their shared multi-modal information experiences from the videos, which helps recreate real-world dialogue context. 
This provides valuable information for understanding the interplay of multi-modal information in human communication.

\subsection{Multi-modal Dialogue Models}
With the outstanding performance of large language models~\cite{chatgpt, wang2021ernie, hoffmann2022training, zhang2022opt, chung2022scaling, touvron2023llama} on text-based dialogue, many studies have explored how to incorporate multi-modal information into dialogue models. Due to the diverse characteristics of multi-modal dialogue datasets, there are significant differences in task definitions and model settings. For example, PhotoChat~\cite{zang2021photochat} propose the task of retrieving relevant images of dialogues. Divter~\cite{sun2022multimodal} takes dialogue context as input and generates textual response and image as visual response.

Most of existing multi-modal dialogue models take image and dialogue context as raw inputs and generate textual responses. Their frameworks can be broadly classified into two categories. The first architecture, such as Flamingo~\cite{alayrac2022flamingo} and BLIP-2~\cite{li2023blip}, aims to incorporate multi-modal information into large language models. They design adapter that bridge the semantic gap between vision encoder and large language model, and use the powerful generative ability of large language model to generate responses. The second approach directly integrates information from different modalities to achieve multi-modal interaction and generate responses. Some works~\cite{han2023champagne, zheng2022mmchat, shuster2021multi, wang2021modeling, yang2021open} concatenate vision and text features and input them into multi-modal fusion networks. Some studies also utilize images or context to introduce intermediate information that aids in response generation. VisAD~\cite{shen2021text} fuses image and text to obtain a multi-modal representation, and then predicts the keywords and retrieves images related to the response. VICTOR~\cite{liu2022think} leverages a caption model to obtain image descriptions and extends concepts in the dialogue using a knowledge base. Maria~\cite{liang2021maria} extracts region features and corresponding concepts from images. We evaluate some baselines on TikTalk and modify Maria by adding audio and external knowledge as inputs, respectively.

\begin{figure*}[t]
    \centering
    \includegraphics[width=0.95\textwidth]{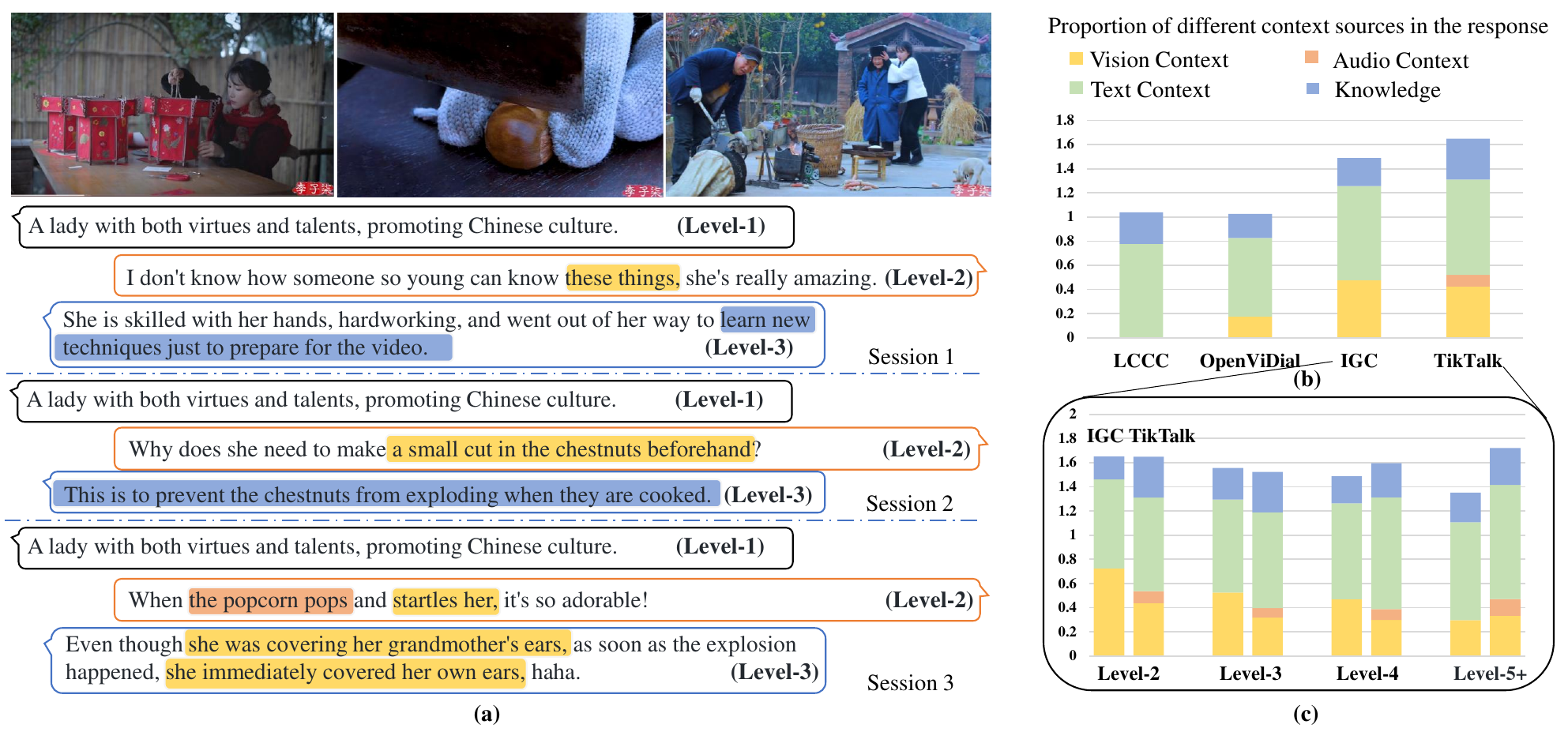}
    \vspace{-0.3cm}
    \caption{Examples and comparisons of responses involving different context sources. (a) Examples of response involving various context sources from TikTalk. We mark in the response the parts related to vision, audio, and external knowledge context with the corresponding colors in (b). (b) Proportions of Responses involving different context sources among four dialogue datasets. (c) Proportions of responses involving different context sources across different levels in IGC and TikTalk.}
    \vspace{-0.3cm}
    \label{fig:teaser}
\end{figure*}

\section{TikTalk Dataset}
\label{sec:data}

We collect data from the Chinese version of TikTok, also known as Douyin, to construct a multi-modal chitchat dataset, which we call \textbf{TikTalk}.
Douyin covers a wide range of video content across more than 25 broad categories, including science, food, gaming, travel, and more. Users have spontaneous chitchat based on the videos, which is similar to the multi-modal chitchat scenes in real world.

\subsection{Data Construction}
We collect videos posted on Douyin in 2021, including their titles, comments, and reply relationships. In order to protect privacy, we don't crawl any user information. 
Not all videos and comments constitute high-quality multi-modal dialogues, so we use some rules to select potentially useful data. First, we remove videos with only one level of comments, as these comments without replies cannot constitute dialogue corpora. Secondly, since users have the option to like videos and comment on the Douyin platform, videos and comments with high numbers of likes are more likely to meet human expectations and have higher quality. We leverage this as a screening criterion and retain videos and comments that meet a certain threshold of likes. Through such filtering, we can obtain high-quality data that is easy to scale.
Furthermore, we clean the comments that we keep. We use regular expressions to filter out useless contents, such as ``@[someone]'' (a way of mentioning another user), merge repeated words, such as ``hahaha'' (laughing with different lengths), and remove some unethical comments. Comments often come with emoticons that can express users' emotions. Therefore, we move emoticons to metadata to keep only the plain text in the dialogues. Finally, we split the hierarchical comments into conversations, forming the final dialogue corpora.

\begin{table}[t]
    \centering
    \caption{Data division of TikTalk. ``Avg-v'', ``Avg-t'', and ``Avg-l'' denote the average duration of videos, the average turns of dialogues, and the average length of utterances.}
    \vspace{-0.3cm}
    \resizebox{\linewidth}{!}{
    \begin{tabular}{lcccccc}
        \hline
          \textbf{Split} & \textbf{Videos} &  \textbf{Dialogues} & \textbf{Turns} & \textbf{Avg-v} & \textbf{Avg-t} & \textbf{Avg-l}\\
        \hline
        \textbf{Train} & 35,703 & 363,589 & 816,170  & 34.01s & 2.24 & 16.42 \\
        \textbf{Val}   & 1,000  & 1,353    & 3,495    & 36.93s & 2.58 & 16.06  \\
        \textbf{Test}  & 2,000  & 2,728    & 7,087    & 36.42s & 2.60 & 16.25 \\
        \textbf{Total} & 38,703 & 367,670 & 826,752 & 34.03s & 2.25 & 16.41 \\
        \hline
    \end{tabular}}
\vspace{-0.5cm}
\label{tab:dataset_statistics}
\end{table}

\subsection{Data Statistics}
Finally, we select 38,703 videos and 367,670 dialogues from a pool of 153,340 videos to construct the TikTalk dataset. The dataset is split into training set, validation set, and test set, consisting of 35,703, 1,000, and 2,000 videos, respectively. The detailed statistics can be found in Table~\ref{tab:dataset_statistics}. These videos have an average duration of 34.03 seconds and provide a variety of multi-modal information. Each dialogue contains an average of 2.25 turns, and due to the transient nature of the chitchat data, responses are more likely to involve video information and external knowledge. To construct high-quality validation and test sets and facilitate evaluation, we select dialogues with at least five genuine responses and chose five of them as the ground-truth responses for that dialogue.

\subsection{Data Analysis}
\label{sec:comparison}
In order to analyze the characteristics of TikTalk, we quantitatively compare the proportions of different context sources involved in the responses in different datasets. Moreover, we define a video-based multi-modal chitchat task and summarize three new challenges.

\subsubsection{Comparison with Other Dialogue Datasets}
We observe that in TikTalk a response cannot be generated or even understood without certain context or knowledge. As shown in the examples in Figure~\ref{fig:teaser} (a), a user would not ask the question ``why does she need to make a small cut in the chestnuts beforehand'' (Level-2 utterance of Session 2) without the vision context of the middle frame. To answer this question, we need to know some external knowledge(Level-3 utterance of Session 2). Sometimes, audio context may be useful, such as in the Level-2 utterance of Session 3, where we need to know that there is a loud sound when the popcorn pops. Therefore, we are wondering whether other datasets have the same phenomena and what the fractions are in different datasets.

We choose three representative dialogue datasets for comparison and randomly pick 300 samples from each dataset:  (1) LCCC~\cite{wang2020large}: textual dialogues collected from social media. There are totally 550 responses in the 300 dialogues; (2) OpenViDial~\cite{meng2020openvidial}: dialogues and aligned key frames extracted from movies and TV series. We obtain 1,019 responses from the 300 sampled key frames; (3) IGC~\cite{mostafazadeh2017image}: image-based chitchat dialogues created by crowdsourcing. There are 1,182 responses associated with the 300 sampled images. For our TikTalk, there are totally 2,472 responses in the 300 sampled videos.
Then, we develop an annotation website and ask annotators to view a video and text context and then judge whether they cannot understand a response if without (1) vision context, (2) audio context, (3) text context, (4) external knowledge. Note that a response can involve context information from multiple modalities. When a response is hard to judge, they can skip labeling it.

The comparison results of the four datasets are shown in Figure~\ref{fig:teaser} (b). Firstly, TikTalk has the highest overall proportion, indicating that it has the richest multi-modal context, including 42\% vision, 10\% audio and 34\% knowledge. Secondly, TikTalk has the highest proportion of responses derived from external knowledge (about 34\%), which suggests that people tend to introduce more new information in chitchat when there is more rich multi-modal context. Thirdly, only 18\% of the responses in the OpenViDial dataset are related to vision context, suggesting that OpenViDial is less suitable for investigating how vision context affects response generation. In contrast, this proportion is 48\% and 42\% in IGC and TikTalk respectively, suggesting that they are more visually oriented. On the other hand, IGC has a higher visual relevance ratio, indicating that its responses rely more on images compared to TikTalk. Furthermore, except for OpenViDial, the proportion of text context is close to 80\% for all three dialogue datasets. This indicates that the dialogue in OpenViDial is less articulated than daily dialogue and requires a greater understanding of the overall storyline.

To further explore how it changes when the level of utterances goes deeper in TikTalk, we compare it with IGC in Figure~\ref{fig:teaser}(c). The results indicate that when the level goes deeper, there are still 35\% utterances are related to vision context and the proportion of audio context also stays similar. This is different from our observation in IGC, where the proportion of vision context has a significant decrease trend when the level goes deeper. This may be caused by the richer information provided by a video (in TikTalk) rather than an image (in IGC). Video users can focus on different points of interest and details in the video, and use them to continue discussions.

\subsubsection{Task Definition and New Challenges}
Based on the above comparisons, We formulate the task of video-based multi-modal chitchat as follows. Suppose that we have a multi-modal dialogue dataset $\mathcal{D}$. Each dialogue is represented as $(C^{v}, C^{a}, C^{t}, R)$, where $C^{v}$ and $C^{a}$ represent the vision context and audio context in the video, $C^{t}$ is the textual context formed by the preceding utterances $\{u_{i}\}_{i=1}^{n}$, and $R$ is a textual response. The goal of multi-modal chitchat is to learn a model $P(R|C^{v},C^{a},C^{t})$ from $\mathcal{D}$ to generate an ideal response $R$ to the given context $C^{v}$, $C^{a}$, and $C^{t}$.

There are three new challenges in addressing this multi-modal dialogue problem on TikTalk:

1) \textbf{Multi-modal Diversity}: Due to the diversity of multi-modal information in TikTalk, users generate diverse responses involving different modalities for a given utterance, e.g., the three Level-2 utterances of different session in Figure~\ref{fig:teaser} (a), which are replied by users to the same Level-1 comment after viewing the video. Without the vision context, the user cannot post the question of why to make a small cut in the chestnuts. Without the audio context, it is difficult for the users to realize the moment when the popcorn pops. Therefore, it is essential to perceive more context information and comprehend the interaction among different information.

2) \textbf{Capturing Interests}: Compared with image-based conversations, it is more challenging to automatically capture points of human interest in video to generate relevant responses. For example, users focus on different points of interest in a video and issue the three Level-2 utterances shown in Figure~\ref{fig:teaser} (a). Despite having all the visual and audio information, we are still uncertain which part should be taken as the point of interest  to evoke the desired response. When models use a single image as input, they can easily take it as the point of interest. However, when the image is extended to
a video, it is non-trivial to decide which video or audio clips should be used as a multi-modal prompt along with a text prompt.  It is also difficult to understand some human interests in video clips. For example, the Level-3 utterance in Session3 involve understanding the change of the lady covering whose ears.

3) \textbf{Complementing Knowledge}: Despite users being unaware of it, they use a lot of knowledge beyond the video when composing a response. For instance, the Level-3 utterance of Session 1 uses the background information of the video creator, Li Ziqi, who is diligent in learning new skills for preparing her videos. For another example, to reply the Level-2 utterance in Session 2, we need the knowledge that making a cut before cooking is to prevent the chestnuts from exploding. Such external knowledge mostly likely can be neither learnt from training data nor generalized for a new video. It is necessary to introduce external knowledge in generating responses somehow. Using large language models may be one way while complementing knowledge graphs may be another.
More detailed statistics and analyses of TikTalk are shown in Appendix ~\ref{sec:dataset details}.

\begin{figure*}[t]
    \centering
\includegraphics[width=0.95\textwidth]{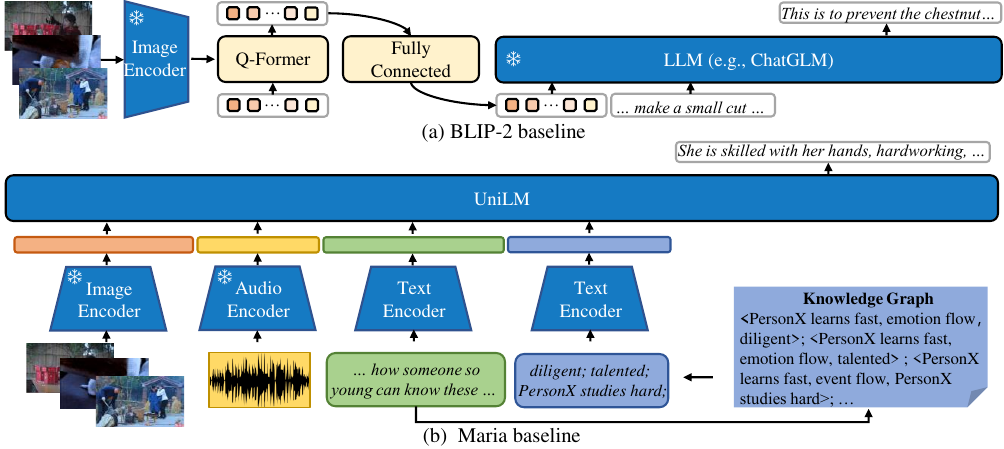}
\vspace{-0.2cm}
    \caption{Two mainstream frameworks for multi-modal dialogue tasks, as exemplified by BLIP-2 and  the modified model based on Maria. (a) These models design adapters (such as Q-Former) to bridge the semantic gap between frozen large language models and pre-trained vision encoders and use large language models to generate responses. (b) The extracted information from various modalities is concatenated directly to perform multi-modal interaction and fusion to generate responses. Based on Maria, we introduce audio information and external knowledge as additional inputs, respectively.}
    \vspace{-0.2cm}
    \label{fig:model_example}
\end{figure*}

\section{Experiments}
\label{sec:exp}

\subsection{Evaluate Metrics}
\subsubsection{Automatic Metrics}
We adopt two types of automatic metrics to evaluate the performance of models from different perspectives:
\begin{itemize}[wide, labelwidth=1pt, labelindent=0pt]
\item \textbf{Similarity}: For each dialogue, we calculate BLEU-2, BLEU-4~\cite{papineni2002bleu}, Meteor~\cite{denkowski2014meteor}, Rouge-L~\cite{lin2004rouge} and CIDEr~\cite{vedantam2015cider}. Each conversation in the validation and test sets has five ground-truth responses. We calculate these metrics for each ground-truth response separately and take the averages as the results for each conversation. We adopt the public NLG evaluation code\footnote{\url{https://github.com/Maluuba/nlg-eval}} to calculate these metrics.

\item \textbf{Diversity}: We adopt Dist-2 and Dist-3~\cite{li2016diversity} to measure the diversity of generated responses.

\end{itemize}
\subsubsection{Human Evaluation} To further evaluate the quality of generated responses, we conduct human evaluation on four metrics:
\begin{itemize}[wide, labelwidth=1pt, labelindent=0pt]
\item \textbf{Sensibleness}: Whether the response is meaningful and coherent, and conforms to logic and common sense.

\item \textbf{Specificity}: Whether the response is relevant to specific context, provides sufficient details, and avoids being too vague or general.
\item \textbf{Multi-modal Relevance}: The degree of correlation between the response and the multi-modal information in the video.
\item \textbf{Overall Quality}: Overall assessment of the quality of the generated response based on the video and dialogue context.
\end{itemize}
Each metric has five levels, ranging from 1 point (worst) to 5 points (best). We randomly select the generation results of 100 dialogues from the test set and ask three annotators to independently evaluate them. The final score is calculated as the mean of their ratings.

\subsection{Baselines and Experimental Setup}
We evaluate some state-of-the-art baselines from different tasks and settings. Existing architectures of multi-modal dialogue models can mainly be classified into two categories, as shown in Figure~\ref{fig:model_example}. Inspired by the findings in Section~\ref{sec:comparison}, we also have trials to modify Maria by adding audio context and knowledge.

\begin{itemize}[wide, labelwidth=1pt, labelindent=0pt]

\item \textbf{DialoGPT}: ~\cite{zhang2019dialogpt} is an extension of GPT-2~\cite{radford2019language} that has been pre-trained on a massive corpus of conversational data. We fine-tune a Chinese version of the DialoGPT model~\footnote{\url{https://github.com/yangjianxin1/GPT2-chitchat}} on TikTalk.

\item \textbf{ChatGLM-6B~\footnote{\url{https://github.com/THUDM/ChatGLM-6B}}}: ~\cite{zeng2023glm-130b} is a conversational large language model based on the GLM architecture~\cite{du2022glm}. It employs similar techniques to ChatGPT.
As it is biased towards receiving input prompts in the form of questions, we fine-tune it on TikTalk, which enables it to generate appropriate responses for non-Q\&A type dialogues. 

\item \textbf{Livebot}: is proposed for live comment generation~\cite{ma2019livebot}. It is based on the Transformer architecture~\cite{vaswani2017attention} and consists of three modules: a video encoder, a context encoder, and a response decoder. 

\item \textbf{BLIP-2}: ~\cite{li2023blip} is a vision-language framework, as shown in Figure~\ref{fig:model_example} (a). It extracts vision features using a frozen pre-trained image encoder, trains a Q-Former to align them with the textual semantic space, and finally uses a frozen large language model to decode the aligned vision features and textual prompts to generate outputs. We employ ViT-G/14 from EVA-CLIP~\cite{fang2022eva} as visual encoder and the fine-tuned ChatGLM-6B as large language model. They are both frozen during the training of Q-Former. We explore two settings: (1) \textbf{BLIP-2\_Img} encodes one randomly sampled frame of the video, which is the same as the original BLIP-2; (2) \textbf{BLIP-2\_Video} extracts multiple frame features and averages them.

\begin{table*}[t]
\centering
\caption{The evaluation results of automatic metrics for different baseline models on TikTalk (above) and the ablation study on Maria (below). We use two types of automatic metrics to evaluate the models: similarity and diversity metrics.}
\vspace{-0.3cm}
\setlength{\tabcolsep}{3.5mm}{
  \begin{tabular}{l|l|ccccc|cc}
    \hline
     Type & Methods & BLEU-2 & BLEU-4& Meteor & Rouge-L& CIDEr & Dist-2 & Dist-3\\
    \hline
    T baselines& DialoGPT & 3.76 & 0.73 & 4.87 & 10.43 & 13.50 & 26.82 & 55.36\\
     & ChatGLM-6B & 3.14 & 0.95 & 3.84 & 9.34 & 16.73 & 36.28 & 58.79 \\
    \hline
    V+T baselines& Livebot  & 0.33 & 0.06 & 1.20 & 4.40 & 5.94 &  0.10 & 0.12\\
     & BLIP-2\_Img & 3.42 & 0.87 & 3.77 & 8.34 & 12.05 & \textbf{38.40} & \textbf{64.63} \\
     & BLIP-2\_Video & 3.41 & 0.75 & 3.72 & 8.11 & 11.61 & 37.32 & 62.44 \\
     & Maria& 4.31 & 1.48 & 5.07 & 11.60 & 26.19 & 29.31 & 48.04\\
    \hline
    V+T+A baseline & Maria+Audio &3.01 & 0.96 & 4.15 & 10.19 & 20.67  & 24.60 & 41.11\\ \hline
    V+T+K baseline  &Maria+C$^3$KG& \textbf{4.74} & \textbf{1.64} & \textbf{5.32}& \textbf{11.84} & \textbf{26.95}  & 32.82 & 54.62\\
    \hline
    \hline
    Maria ablations & text & 3.35 & 1.18 & 4.56 & 11.16 & 24.90  & 26.83 & 43.80 \\
    & Audio+text & 2.92 & 0.95 & 4.13 & 10.19 & 20.52  & 23.95 & 39.78\\
    & C$^3$KG+text & 4.34 & 1.56 & 5.09 & 11.65 & 26.78  & 31.24 & 51.38\\

    \hline
\end{tabular}}
\vspace{-0.3cm}
\label{tab:baseline_performance}
\end{table*}

\item \textbf{Maria}: is a visual conversation model~\cite{liang2021maria} that employs the UniLM architecture~\cite{dong2019unified} for multi-modal modeling. We extract consecutive video frames as visual inputs and use MCP (Masked Context Prediction) and MRP (Masked Response Prediction) as training objectives. In addition, we make two modifications to Maria, which involve introducing audio and external knowledge as inputs to Maria, respectively, as shown in Figure~\ref{fig:model_example} (b).
\textbf{Maria+Audio} uses the VGGish network~\cite{hershey2017cnn} to extract audio features from the video. Then we take consecutive audio frames as another segment as video frames do.
\textbf{Maria+C3KG} retrieves dialogue flows from a commonsense dialogue knowledge base as implicit context. As we observe in Figure~\ref{fig:teaser}, about 34\% of the responses in TikTalk are related to external knowledge. We use C$^3$KG~\cite{li2022c3kg}, a Chinese commonsense conversational knowledge graph. Based on people's emotions and reactions to daily events, C$^3$KG constructs three types of dialogue flows, including event, concept, and emotion-based empathy flow. For each conversation, we use Sentence-BERT ~\cite{reimers2019sentence} to encode textual context. By measuring the semantic similarity, we retrieve the heads of dialogue flow in C$^3$KG which are most relevant to the dialogue context. Next, we obtain the corresponding commonsense conversational knowledge and employ it as knowledge input. In addition, we add MKP (Masked Knowledge Prediction) as a training objective. We use a probability of 15\% to mask textual context and knowledge, while in response prediction the probability is 70\%. The implementation details are shown in Appendix~\ref{sec:implementationt details}.

\end{itemize}

\subsection{Main Results}
Experimental results for the automatic metrics are presented in the upper part of Table~\ref{tab:baseline_performance}. The modified Maria+C$^3$KG model outperforms the other baselines in terms of similarity metrics, and improves over Maria in all metrics. This validates the usefulness of introducing knowledge as implicit context in multi-modal dialogue. While the performance of Maria+Audio was inferior to that of Maria.
There are three possible reasons for this phenomenon. Firstly, as depicted in Figure~\ref{fig:teaser} (b), the proportion of responses involving audio context is less than 7\%, in contrast to over 40\% for visual context and over 30\% for knowledge context. The available data may not be sufficient to adequately train the model to effectively utilize audio context. Secondly, unlike the text modality, the audio modality is low-level and challenging to extract the high-level information relevant to our problem. Consequently, the model may not effectively learn how to utilize audio context. In our experiment, we employ VGGish as a feature extractor, which is inclined to extract ambient sound information rather than speech information. Thirdly, given that the majority of cases heavily depends on visual and textual information, the inclusion of low-level and noisy audio tokens has the potential to disrupt the original generation models.
Due to the rich knowledge contained in large language models, the models that utilize ChatGLM-6B perform exceptionally well on diversity metrics. Especially, BLIP-2 build upon this by incorporating visual information from videos, further enhancing the diversity of generated responses. The performance of Livebot is poor on all metrics, possibly due to its design for generating live comments, leading to the generation of numerous repetitive and similar sentences.

The human evaluation results in Table~\ref{tab:human_evaluation} show that Maria+C$^3$KG performs the best in terms of sensibleness, multi-modal relevance, and overall quality, while BLIP-2\_Video achieves the highest score in specificity. Maria+C$^3$KG outperforms Maria on all four metrics, while the performance of Maria+Audio decreases, which is consistent with the results of automatic metrics. ChatGLM-6B outperforms DialoGPT thanks to its larger pre-training corpus and the training process similar to ChatGPT, which enable such large language models to possess inherent knowledge. BLIP-2\_Video performs better than BLIP-2\_Img, and we attribute this to the fact that it takes into account the entire video information. Their responses are more specific compared to ChatGLM-6B, possibly because they streamline the visual information and have clearer focuses.

\begin{table}[t]
\centering
\caption{Results of human evaluation. We evaluate the responses on sensibility(Sensible.), specificity(Specific.), multi-modal relevance(MM Relevant.), and overall quality(Quality).}
\vspace{-0.25cm}
\setlength{\tabcolsep}{1.1mm}{
  \begin{tabular}{lcccc}
    \hline
     Methods & \textbf{Sensible.} & \textbf{Specific.}& \textbf{MM Relevant.} & \textbf{Quality}\\
    \hline
    DialoGPT & 2.37 & 1.91 & 1.73 & 2.00\\
    ChatGLM-6B & 2.87 & 2.12 & 2.07 & 2.36 \\
    Livebot & 2.29 & 1.03 & 1.68 & 1.68 \\
    BLIP-2\_Img & 2.66 & 2.25 & 2.03 & 2.33 \\
    BLIP-2\_Video & 2.79 & \textbf{2.43} & 2.14 & 2.48 \\
    Maria & 3.06 & 2.27 &  2.15 & 2.53 \\
    Maria+Audio & 2.98 & 2.17 & 2.15 & 2.45\\
    Maria+C$^3$KG & \textbf{3.29} & 2.41 & \textbf{2.31} & \textbf{2.71}\\
    \hline
\end{tabular}}
\vspace{-0.45cm}
\label{tab:human_evaluation}
\end{table}

\begin{figure*}[t]
    \centering
    \includegraphics[width=\textwidth]{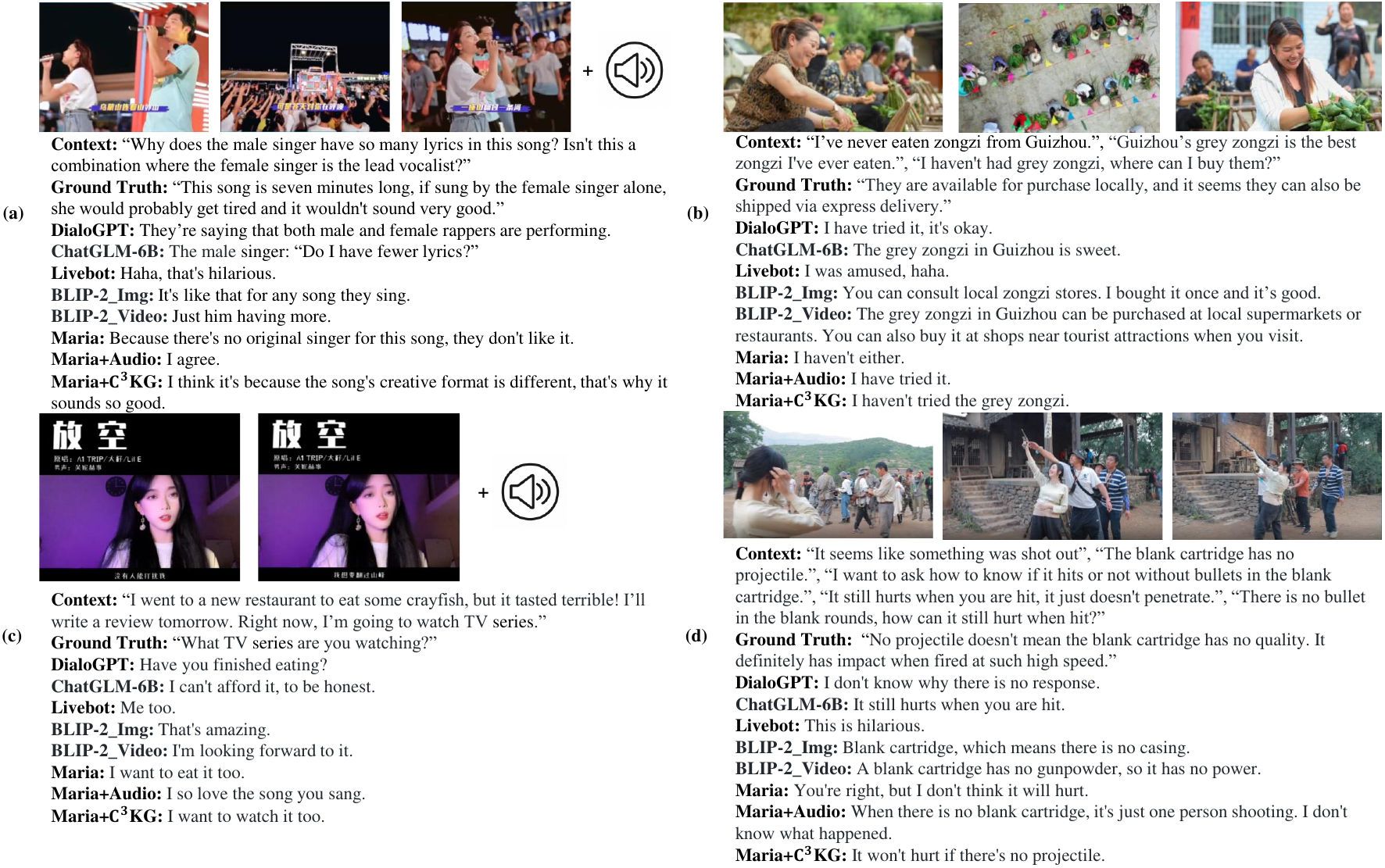}
    \vspace{-0.7cm}
    \caption{Examples of generated response from models. (a) Two singers are singing on the stage. (b) Some people are making zongzi. (c) A video blogger is recording a new song. (d) An actress is learning how to use a prop gun from the crew. }
    \vspace{-0.4cm}
    \label{fig:case_study}
\end{figure*}

\subsection{Ablation Study}

The results of our ablation study are shown in the last three rows of Table~\ref{tab:baseline_performance}. We evaluate Maria, Maria+Audio, and Maria+C$^3$KG by removing visual information. We observe decrease in the results of the three models, which shows that the visual context enhance the quality of generated responses. In contrast, the model that combines audio and text performs worse than the one that only includes text, suggesting that simply fusing audio features is not a viable solution. On the other hand, the Maria+C$^3$KG model without visual information, i.e. C$^3$KG+text, shows the smallest degradation among the three ablations, and even slightly outperforms Maria. This suggests that external knowledge is equally important as visual context, which is consistent with the analysis results in Section~\ref{sec:comparison}.

\subsection{Case Study}
To further investigate the performance of the baselines, we present some response cases in Figure~\ref{fig:case_study}. In Figure~\ref{fig:case_study}(a), Maria+C$^3$KG is able to infer the reason for the lyric allocation. Figure~\ref{fig:case_study}(b) shows that the baselines using the large language model can provide responses based on common sense. Impressively, the response from Maria+Audio contains audio information that is not mentioned in the textual context in Figure~\ref{fig:case_study}(c). These three cases illustrate that the external knowledge and the audio information all play a role in multi-modal dialogue. However, as shown in Figure~\ref{fig:case_study}(d), for more difficult logical reasoning problems, these models may be powerless and produce unreasonable responses. And no model can produce satisfactory responses in all these cases. In summary, due to the diversity of video scenarios, TikTalk requires complex perception, understanding, and reasoning capabilities. While existing baselines can sometimes produce reasonable responses, they fall far short of meeting the expectation of multi-modal dialogue in the real world.

\section{Conclusion}
\label{sec:conclusion}

In this paper, we introduce TikTalk, a video-based multi-modal chitchat dataset. We collect videos from a video-sharing platform and users' spontaneous dialogues based on them. We analyze the characteristics of TikTalk through quantitative comparisons with other dialogue datasets. Additionally, we propose a multi-modal chitchat task and summarize three new challenges posed by TikTalk, including perceiving and understanding more diverse multi-modal context and information interaction, capturing the human interests from rich information, and supplementing with external knowledge. We evaluate some baselines and experimental results show that existing methods exhibit some advantages, but they cannot address the above challenges well and generate ideal responses. There still remains a considerable gap between the performance of multi-modal dialogue models and humans, indicating great potential for further improvements on TikTalk. We hope this work can contribute to the research on multi-modal dialogue.

\section{Limitations and Future Work}
\label{sec:discussion}

For the new challenges presented by TikTalk, existing baselines have demonstrated some abilities but still have limitations. Firstly, simple fusion of audio information does not lead to improved model performance. Experimental results indicate that we need better methods to leverage audio in videos, including sound and transcripts in text. Secondly, current models rely on attention mechanisms to capture the points of human interest in videos, while such implicit way may not be optimal. In the future, we may try ground utterances with frames or regions in frames to explicitly supervise our algorithm to automatically capture human interests. Third, knowledge is surprisingly helpful. Although the BLIP-2 based on LLM performs worse than Maria integrated by a knowledge graph, it is still open whether the BLIP-2 model based on LLM can leverage the knowledge graph too. Actually they may excel in different knowledge domains. LLM knows why the chestnuts shall be made a cut before cooking while the knowledge graph that we use knows better that a person may possess many skills due to one's diligence. The flows in C$^3$KG is like a chain-of-thoughts (CoT) in the emotion, reaction, and event domains. If we can enable LLM the ability of CoT in these domains somehow, perhaps the BLIP-2 based on LLM can think step by step and infer the appropriate response.

%%
%% The acknowledgments section is defined using the "acks" environment
%% (and NOT an unnumbered section). This ensures the proper
%% identification of the section in the article metadata, and the
%% consistent spelling of the heading.

\begin{acks}
This work is supported by the Fundamental Research Funds for the Central Universities, and the Research Funds of Renmin University of China (21XNLG28) and National Natural Science Foundation of China (No. 62276268).
\end{acks}

%%
%% The next two lines define the bibliography style to be used, and
%% the bibliography file.
\bibliographystyle{ACM-Reference-Format}
\balance
\bibliography{sample-base}

%%
%% If your work has an appendix, this is the place to put it.
\clearpage
\appendix

\section{Statistical Details of TikTalk}\label{sec:dataset details}
\subsection{Statistics of Topics}
We gather topic category names from Douyin and assign videos to these categories. Actually, we employ Sentence-BERT to calculate the similarity between their descriptions and the topic category names. We show the number of videos in each category in table ~\ref{tab:category statistics}.

\begin{table}[htp]
    \centering
    \caption{Statistics of videos in different Topics on TikTalk.}
    \resizebox{\linewidth}{!}{
    \begin{tabular}{cccccc}
        \hline
         Topic &	Science &	Short play	&Gaming	&Lifestyle&	Recording	\\
         Videos	&8448	&4684	&4265&	3592&	2487\\
         \hline
        &Food	&Creativity	&Parenting	&Campus	&Others\\
         &	1973&	1660	&1596	&822	&9176 \\
        \hline
    \end{tabular}}

\label{tab:category statistics}
\end{table}

\subsection{Distribution of Human Interests}
Human interests encompass various aspects of a video that capture users' attention, prompting them to respond based on their specific points of interest. For instance, as depicted in Figure~\ref{fig:teaser}, the three Level-1 comments illustrate distinct user focuses: one user directs attention to the video clip demonstrating the process of cutting chestnuts, while another expresses interest in the auditory and visual experience of popcorn exploding. Consequently, human interests involve multiple modalities and necessitate a nuanced comprehension that evolves over time.

We analyze the human interests in visual modality. Specifically, we conduct preliminary statistical analysis to examine the correlation between visual information at various time segments and human responses. Utilizing the Chinese-CLIP~\cite{yang2022chinese} model, we measure the similarity between responses annotated as relevant to the visual context and all video frames. Subsequently, we identify the time segment in the video that contains the most similar frame. We then perform statistical analysis across different time segment buckets and present the results in table~\ref{tab:distribution of human interests}. It indicates that users display relatively uniform interests over time in responses related to the visual context.

\begin{table}[htp]
    \centering
    \caption{The distribution of human interests in the visual modality across different time segments in videos.}
    \resizebox{\linewidth}{!}{
    \begin{tabular}{cccccc}
        \hline
         Time Segment (\%)&	0\textasciitilde20&	20\textasciitilde40	&40\textasciitilde60	&60\textasciitilde80	&80\textasciitilde100\\
         \hline
         Relevant 	&224&	172	&229&	194&	227 \\
         Responses&(21.4\%)&	(16.4\%)	&(21.9\%)&	(18.6\%)&	(21.7\%) \\
        \hline
    \end{tabular}}

\label{tab:distribution of human interests}
\end{table}

\subsection{Classification of External Knowledge}
We classify external knowledge into two primary categories: general knowledge and personalized knowledge. For instance, in Session 2 of Figure~\ref{fig:teaser}, the Level-3 commentator provides an explanation regarding the necessity of making a small cut in chestnuts before cooking, which falls under general knowledge. On the other hand, personalized knowledge encompasses information pertaining to the video blogger, as well as the personal experiences and preferences of users. For instance, in Session 1 of Figure~\ref{fig:teaser}, the Level-3 commentator demonstrates awareness of the video blogger's background and highlights the extensive preparations made in advance for the video, explaining this to the Level-2 commentator. Users may watch other videos by the same blogger to acquire additional knowledge, thereby combining personalized information to formulate their responses. To evaluate these categories, we randomly assign 100 responses related to external knowledge annotated in Figure~\ref{fig:teaser} (b). The results indicate that 72 out of 100 examples are classified as general knowledge, while the remaining 28 examples are associated with personalized knowledge.

\section{Implementationt Details}\label{sec:implementationt details}
In our implementation of Maria, we adopt ResNet101~\cite{he2016deep} to extract the video frame features. The maximum length of the input sequence is 180 textual tokens and 50 frames for both vision and audio. We initialize the parameters with a pre-trained model Bert-base-Chinese and train 20 epochs with a linear decay of learning rate from $3\times 10^{-5}$.

We conduct statistical analysis of the training efficiency of the Maria models using various modal combinations, employing 4 Nvidia V100 (32G) GPUs. The results are presented in table~\ref{tab:modality efficiency}. When incorporating an additional modal input for Maria, the training time increases by 37\% to 75\%. Despite this added cost, it is deemed acceptable due to the substantial improvement in generation quality achieved by combining certain modalities. Inference efficiency is also acceptable as online inference can benefit from acceleration strategies such as distributed processing, caching, and compression. In our future work, we will explore methods to expedite and optimize model training and inference.

Regarding the human evaluation, we calculate the Fleiss' Kappa, which yields a value of 0.52 (-1\textasciitilde1), indicating a substantial agreement among the three annotators. We also try a small-scale test with more annotators, while the results are very similar to our current annotations.

\begin{table}[htp]
    \centering
    \caption{Traing efficiency of different combination of modalities. Text (T), Image (I), Audio (A), Knowledge (K).}

    \begin{tabular}{lc}
        \hline
        Modalities & Training Efficiency(min/epoch) \\
        \hline
        M1: T & 10 \\
        M2: T+I & 17.5 (vs. M1, +75\%) \\
        M3: T+I+A & 24 (vs. M2, +37\%)	\\
        M4: T+I+K & 30 (vs. M2, +71\%) \\
        M5: T+I+A+K & 37 (vs. M3, +54\%) \\

        \hline
    \end{tabular}

\label{tab:modality efficiency}
\end{table}

\end{document}